\documentclass[lettersize,journal]{IEEEtran}
\usepackage{amsmath,amsfonts}
\usepackage{algorithmic}

\usepackage{array}
\usepackage[caption=false,font=normalsize,labelfont=sf,textfont=sf]{subfig}
\usepackage{textcomp}
\usepackage{stfloats}
\usepackage{url}
\usepackage{verbatim}
\usepackage{graphicx}
\usepackage{cite}

\usepackage[ruled]{algorithm2e}

\newtheorem{myDef}{\textbf{Definition}}

\begin{document}

\title{FedSiam-DA: Dual-aggregated  Federated Learning via Siamese Network under Non-IID Data}

\author{Ming Yang,~\IEEEmembership{Member,~IEEE,} Yanhan Wang, Xin Wang,~\IEEEmembership{Member,~IEEE,} Zhenyong Zhang,~\IEEEmembership{Member,~IEEE,} Xiaoming Wu, Peng Cheng,~\IEEEmembership{Member,~IEEE,}

\thanks{M. Yang Y. Wang, X. Wang, and X. Wu are with Shandong Provincial Key Laboratory of Computer Networks, Shandong Computer Science Center, Qilu University of Technology (Shandong Academy of Sciences), Jinan 250014, P. R. China. Emails: {\small yangm@sdas.org, yanhanwang.ww@gmail.com, xinw.zju@gmail.com, wuxm@sdas.org}}

\thanks{Z. Zhang is with  the State Key Laboratory of Public Big Data, College of Computer Science and Technology, Guizhou University, Guiyang 550000, China. Email: {\small zyzhangnew@gmail.com }}
\thanks{P. Cheng is with the State Key Lab. of Industrial Control Technology, Zhejiang University, Hangzhou 310027, P. R. China. Email:
	{\small lunarheart@zju.edu.cn}}
}

\maketitle

\begin{abstract}
Federated learning is a  distributed learning that allows each client to keep the original data locally and only upload the parameters of the local model to the server. Despite federated learning can address data island, it remains challenging to train with data heterogeneous in a real application. 
In this paper, we propose  FedSiam-DA, a novel dual-aggregated contrastive federated learning approach,  to personalize both local and global models, under various settings of data heterogeneity.  
Firstly, based on the idea of contrastive learning in the Siamese Network, FedSiam-DA regards the local and global model as different branches of the Siamese Network during the local training and controls the update direction of the model by constantly changing model similarity to personalize the local model.  
Secondly,  FedSiam-DA introduces dynamic weights based on model similarity for each local model and exercises the dual-aggregated mechanism to further improve the generalization of the global model.
Moreover, we provide extensive experiments on benchmark datasets, the results demonstrate that FedSiam-DA achieves outperforming several previous FL approaches on heterogeneous datasets. 
\end{abstract}


\section{Introduction}
With the rapid increase in computing power and storage capacity of the devices in modern distributed networks,  a wealth of local resources is generated. Although the abundance of data provides great opportunities for the application of artificial intelligence, the direct collection of raw data faces many problems\cite{mao2017survey,abbas2017mobile,zhang2022elastic}. Firstly, due to privacy concerns and data security regulations, many users, especially government facilities, medical institutions, and financial institutions, choose to keep data locally, which has led to the emergence of data islands\cite{kairouz2021advances,jiang2020federated}. Besides, large numbers of data uploads brings huge communication overhead\cite{mcmahan2017communication}.

Federated learning allows each client to stay the original data locally and only upload the parameters updated by the local model, which avoids data sharing between clients and is an effective method for data privacy protection \cite{wang2020privacy,wang2021dynamic} in the current artificial intelligence context \cite{lim2020federated,li2021survey,li2020federated}. 
It aims at aggregating local models from each client to get a global model which can achieve more robustness and general.
Federated learning allows multiple clients to collaboratively train a shared global model, whose performance is highly dependent on the data distribution among clients.
Therefore, traditional federated learning is usually carried out under the assumption that the data of each client is Independent Identically Distribution (IID)  \cite{yang2019federated,li2019convergence}, which is more suitable for scenarios where clients are similar to each other in their private data distribution.

However, in many application scenarios, data heterogeneity is a common phenomenon for different clients. Due to the data being Non-Independent Identically Distribution (Non-IID) \cite{li2022federated,zhang2021delay,wang2018privacy,sattler2019robust,xiong2021privacy} on each client, each local model update in the direction of the local optimum, leading to a large audience in the update direction of different local models and deviating from the optimal global model, resulting in the phenomenon of client drift\cite{li2021decentralized,mitra2021linear}.
Furthermore, the more local updates, the greater distance between each local model.
At the same time, if the server aggregates these local models to obtain the global model directly, which will also deviate significantly from the optimal global model\cite{deng2020adaptive}.  
Moreover, sharing this global model with all clients may lead to poor performance or slow convergence of the local model.
Therefore, the global model obtained by directly averaging and aggregating each local model cannot fit each local dataset well, which has poor generalization and fairness on the heterogeneous dataset.

Carrying the above insight, in this paper, we address the data heterogeneity challenge by optimizing the federated learning algorithm on both clients and servers. 
On the client, we strive to control the update direction\cite{li2022federated,shamsian2021personalized} of each local model to reduce the distance between them, thereby alleviating the phenomenon of model skew.
On the server, we personalize global models by adjusting aggregate weights \cite{yu2019federated} to fully learn knowledge from heterogeneous datasets. We introduce dynamic weights to further improve the generalization ability of the global model and make it perform more equitably on different heterogeneous datasets.

Siamese Network\cite{chicco2021siamese,dong2018triplet} converts the inputs of the left and right neural networks into a feature vector respectively. Then, it calculates the similarity of the two inputs in the new feature space, thereby minimizing the distance between the two inputs. 
Inspired by the siamese network, changing the similarity between models over heterogeneous datasets can effectively correct the direction of model updates, thereby reducing the gap of each local model to mitigate the effect of data heterogeneity on federated learning.

In this paper, we propose a novel Dual-aggregated Contrastive Federated Learning approach based on the siamese network (FedSiam-DA) which adopts the stop-gradient mechanism to alternate the distance between models.
As we all know, due to data heterogeneity leads to inconsistent update directions of each local model, they can only be applied to their local dataset, and the generalization is poor. Nevertheless, the global model obtained by aggregating each local model has better generalization than the local model. 
Therefore, we strive to control the updated direction of the current local model towards the global model and away from the local model of the previous round during the local training. Accordingly, each local model can better learn the generalization of the global model and not negatively affect heterogeneous data in the update direction.
Furthermore, we introduce a stop-gradient\cite{chen2021exploring,tao2022exploring} mechanism to alternately optimize the global and local models by adversarial training. 
Therefore, the local model can increase its similarity with the global model while achieving better performance on its local dataset.
On the server, we personalize the global model with a dual-aggregated mechanism. In the first aggregation, we follow the aggregation method of FedAvg \cite{mcmahan2017communication}, and the server obtains the parameters of the global model by the weighted average of the received local model parameters. In the second aggregation, we calculate the cosine similarity between the local and global model obtained in the first aggregation to redefine the weight of each local model which participates in the second aggregation. Introducing dynamic weights for each local model so that the global model can better learn the knowledge of each local model and fully mine the value of heterogeneous data.

Our main contributions can be summarized as follows:
\begin{itemize}
	\item We optimize federated learning at both the client and server sides, which brings a double optimization effect. 
	\item In local training, we modify the update direction of the local model by controlling the cosine similarity between the output of models.
	\item We introduce a stop-gradient mechanism to alternately optimize global and local models in an adversarial training manner during local training.
	\item  For global aggregation, we perform dual-aggregated on the server, and design a dynamic weight for each local model in the second aggregation to further improve the generalization of the global model.
	\item Extensive experiments show the excellence of our proposed method in terms of test performance in a data-heterogeneous environment constructed using several benchmark datasets.
\end{itemize}

\section{Related Work}
\subsection{Heterogeneous Federated Learning}
Recently, federated learning has become a hot research topic as an effective method of data privacy protection. As a classic FL framework, FedAvg \cite{mcmahan2017communication}  manages weighted parameter averaging to update the global model. FedAvg achieves asymptotic convergence when the data is IID on each client. However, FedAvg convergence degrades significantly for clients with heterogeneous data. The studies in \cite{li2019convergence,khaled2020tighter} demonstrate that client drift during local updates caused by the Non-IID distribution of each client data is the main reason that ultimately leads to the degradation of convergence rates.
Previous work has shown that data heterogeneity introduces challenges to federated learning, such as client drift, model skew, and slow convergence. 
Moreover, the global model obtained by averaging and aggregating local model parameters is difficult to apply to different clients which hold Non-IID distributed data.
Therefore, some works try to personalize federated learning mechanisms\cite{tan2022towards,zhu2021federated} from the perspectives of local training and global aggregation, to improve the convergence rate of federated learning when the data of each client is heterogeneous.\\[2pt]

\noindent\textbf{Local Client Training:}

FedProx\cite{sahu2018convergence} controls the local model to update in the direction closer to the global model by adding a proximal term to the local loss function, in this manner it improves model skew on a different client.
Furthermore, FedProx defines a $\gamma$-inexact solution to dynamically adjust the number of local iterations by imprecisely solving the local function. 
MOON\cite{li2021model} corrects the update direction through model-contrastive loss. The loss in local training consists of loss generated according to the label, such as cross-entropy loss, and model-contrastive loss. Moreover, MOON applies a hyper-parameter $\mu$ to control the weight of the model-contrastive loss term to alter the convergence rate of the model. 
SCAFFOLD \cite{karimireddy2020scaffold} adopts a control variable to correct the direction of system training. There are two methods to update control variables: 1) using local gradient updates; 2) updating according to the difference between the global and local models.
The local control variables contain gradient information that affects the updated direction of the local model.
The aggregated global control variable contains model update direction information for all other clients. 
SCAFFOLD  overcomes gradient differences between local models by introducing control variables during local updates to alleviate client drift.
FedDC \cite{gao2022feddc} employs the learned local drift variables to achieve consistency at the parameter level. The objective function of the client consists of a standard loss function, constraint penalty term, and gradient correction term. The constraint penalty term includes global parameters, the relationship between drift variables and local parameters, and a gradient correction term. Thus, FedDC reduces gradient drift by a constraint penalty term in each training round. 

The global model aggregates multiple local models, and the generalization is better than the local models. Therefore, increasing the similarity between the global and local models to modify the updated direction of the local model during local training is a common idea in existing work. However, these methods ignore the fact that the local optimal model is different from the global optimal model. Simply increasing the similarity between the global and local models will cause each local model to over-learn the generalization of the global model and lose its individuality\cite{huang2022learn,gao2022feddc}. Inability to quickly converge to a local optimum on a local dataset.\\[2pt]

\noindent\textbf{Global Server Aggregation:}

FedAMP\cite{huang2021personalized} considers that data are Non-IID in a real application, a single global model cannot meet all clients' datasets. Therefore, each client has a personalized cloud global model on the server in FedAMP. Hence, the cloud server uses the attentive message-passing mechanism to aggregate the local personalized model and obtain the cloud personalized model for each client.
FedFTG \cite{zhang2022fine} samples each class according to the distribution of all training data in each round.  At the same time, according to the distribution of each class of each client, a class-level integration method is proposed to allocate the aggregation weight of each client's local model when aggregating on the server. 
FedMA \cite{xie2019asynchronous} is a  hierarchical federated learning algorithm based on   PFNM \cite{yurochkin2019bayesian} matching, which requires that the number of communication rounds equals the number of layers in the network, and clients upload weights one layer at a time. The server accomplishes single-layer matching to obtain the weight of the first layer of the global model and broadcasts these to the client, who freezes the matched layers and continues to train all consecutive layers on its dataset until the last layer. 

The above two classes of optimization strategies partially optimize federated learning from the client or server perspective, respectively. Our framework aims to alleviate the problem of slow convergence of federated learning caused by data heterogeneity on both the client and the server side and achieve the effect of double optimization.

\begin{figure*}
	\centering
	\includegraphics[width=0.7\linewidth]{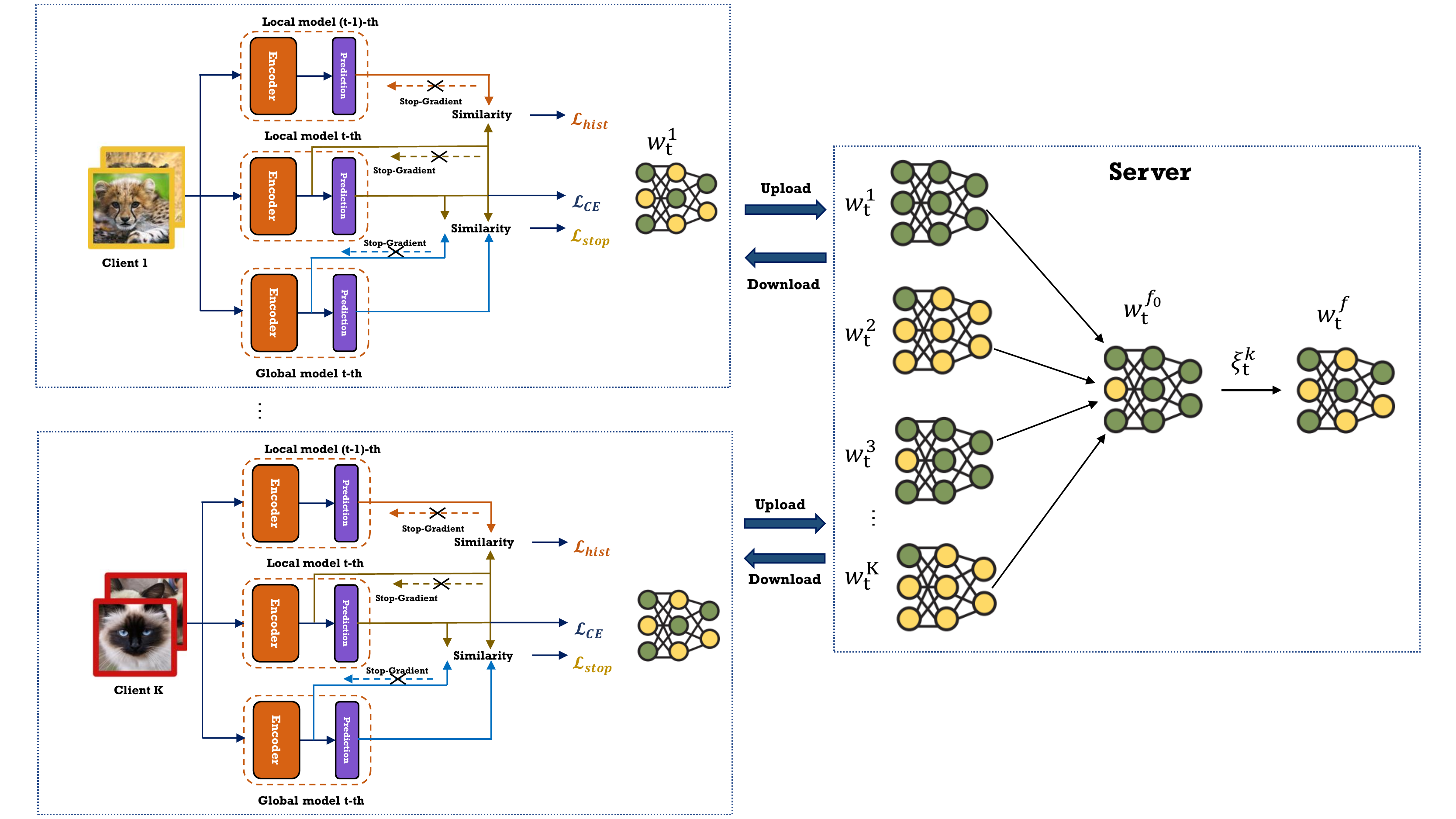}
	\caption{A schematic illustration of our proposed FedSiam-DA framework for model personalization in federated learning. We  introduce  stop gradient mechanism on the client  and apply dual-aggregated on the server.}
	\label{fig:fedsiam-da}
\end{figure*}

\subsection{Siamese Network}
Siamese network first appeared in 1993 to verify that the signature on a check matches the signature retained by the bank, and then mainly applied to compare the similarity of two element vectors in computational fields.
Siamese network includes two networks that share weights to compute comparable output vectors by working together on two different input vectors simultaneously. 
Typically, one of the output vectors is precomputed, thus forming a baseline, to which the other output vector is compared. 
Siamese network  \cite{bromley1993signature,taigman2014deepface} has become a common structure for unsupervised visual representation learning. However, the siamese network suffers from the problem that all outputs "collapse" to a constant. Contrastive learning is a general strategy to prevent the collapse of siamese network.
The key idea of contrastive learning is to decrease the distance between positive sample pairs and increase the distance between negative sample pairs. 
Firstly, the siamese network feeds two inputs into two neural networks to extract the representation vectors of two inputs and map the representations to a latent space. Then,  based on the idea of contrastive learning \cite{hadsell2006dimensionality,wu2018unsupervised}, it maximizes the similarity between the representations of the two inputs.

SimSiam\cite{chen2021exploring} is a novel siamese network framework to improve the similarity between two augmented views of an image.
Moreover,  it is an effective solution to the model collapse problem.
First, SimSiam takes two random augmented views of an image as inputs $x_1$ and $x_2$ to both branches and processes them by the same encoder network $f$ that consists of a backbone and a projection MLP head.
Then, it applies a prediction MLP head $h$ on one side and the stop-gradient mechanism on the other side.                                                                                                                                                                                                                                                                                                                                                                                                                                                                                                                                                                                                                                                                                                                
To maximize the similarity between two inputs, a  symmetrized loss is defined as:                            

\begin{equation}\label{SimSiam}
	\mathcal{L}=\frac{1}{2}D(p_1,stopgrad(z_2))+\frac{1}{2}D(p_2,stopgrad(z_1)),	
\end{equation}\\
where $p_1 = h(f(x_1))$,  $z_2 = f(x_2)$. 
In this loss function, $z_1$ and $z_2$ is treated as constants. 
This means that the encoder on $x_1$ can receive gradients from $p_1$, but the encoder on $x_2$ cannot receive gradients from $x_2$, in the first term. The second term is the opposite process.

In addition to SimSiam, common frameworks based on contrastive learning include MoCo\cite{he2020momentum}, SimCLR\cite{chen2020simple}, BYOL\cite{grill2020bootstrap}, and SimCSE\cite{gao2021simcse}. MOON combines SimCLR with federated learning and proposes model contrastive learning to compare the learned representations of different models. During local training, MOON controls the current local model close to the global model and away from the historical local model. However, the local optimal model is not the same as the global optimal model. Therefore, simply increasing the similarity between the global model and the local model leads to each local model to over-learn the generalization of the global model and losing personality. Inability to quickly converge to a local optimum on a local dataset.
In this paper, we optimize federated learning on both the client and server sides. We append a prediction head MLP above the encoder and introduce the stop-gradient mechanism. On the client side, the global and current local models are optimized alternately by adversarial training to personalize the local model. On the server side, we design a dual-aggregated mechanism to personalize the global model and employ a dynamic aggregation weight for each local model, so that the global model can better learn the knowledge of each local model when the data is heterogeneous.

\section{Personalized Federated Learning}
As show in fig.\ref{fig:fedsiam-da}, each client net consists of two components: a base encoder network $f$ which includes a backbone network (e.g.ResNet \cite{he2016deep}) and a projection MLP head \cite{chen2020simple}, and a prediction MLP $h$ \cite{grill2020bootstrap}.  The base backbone extracts representation vectors from input \textit{x} and the projection MLP \textit{h} map the representations to a latent space, respectively. We, especially, add a prediction MLP head which has 2 layers above the base encoder net to transform the output of one side and match it to the other side.

\begin{algorithm}[] 
	\caption{The FedSiam-DA framework}
	\LinesNumbered 
	\KwIn{number of communication rounds $T$,
		number of clients $N$, number of local
		epochs $M$, learning rate $lr$,
		hyper-parameter $\mu$}
	\KwOut{The final model $w^T$}
	\textbf{Server executes:}\\ 
	\For{$t=0,1...,T-1$}{\
		\For{$k=0,1...,K$ \textbf{in parallel}}{
		send the global model $w^f_t$ to $D^k$\;
		$w_{Mt}^k$ $\gets$ \textbf{LocalTraining}$(k,w^f_t)$
		}
	\textbf{First Aggregation:}\\[3pt]
	 $w_t^{f_0}=\frac{1}{K}\sum_{k=1}^{K}w_t^k$\\[3pt]
	\textbf{Second Aggregation:}\\[3pt]
	$w_t^f=\sum_{k=1}^{K}\xi_t^k w_t^k$	
	}
    return $w^T$\\
    \textbf{LocalUpdate:}\\

	\For{epoch $m = 1, 2, ..., M$}{
		\For{each batch $b =$ (x,y) of $D^k$}{
			$\mathcal{L}_{ce}(w^k_{Mt+m})$ $\gets$ Cross Entropy Loss Function\\[6pt]
			$\mathcal{L}_{stop}=$\\
		\qquad\qquad$\frac{1}{2}D(p_{w^{f'}_{t+m}},stopgrad(z_{Mt+m}^k))$\\[6pt]
			\qquad\qquad $+\frac{1}{2} D(p_{w^k_{Mt+m}},stopgrad(z_{t+m}^{f'}))$\\[6pt]
			
			$\mathcal{L}_{hist}=D(stopgrad(z^k_{Mt+m-1}),z^k_{Mt+m})$\\[6pt]
		
			$\mathcal{L}^k_{Mt+m}=\mathcal{L}_{ce}+\mu (\mathcal{L}_{hist}+ \mathcal{L}_{stop})$\\[6pt]
			
		$w_{Mt+m+1}^k \gets w_{Mt+m+1}^k - lr\nabla\mathcal{L}^k_{Mt+m}$	
		}
	
	}
return $w_{M(t+1)}^k$ to server

\end{algorithm}

\subsection{Local Training}

\noindent\textbf{Problem Statement}

In federated learning, consider $K$ clients $C^1,C^2,...,C^K$, $k \in [K]$ that have a local privacy datasets $D^1,D^2,...,D^K$. These datasets are Non-IID, that drawn from distribution $p^1,p^2,...,p^K$ and each has a personalized local model $w^1,w^2,...,w^K$. 
Meanwhile, $w^{k^*}$ denotes the best performance of $w^k$ can achieve on the distribution $p^k$.
In heterogeneous environments,  local training aims to train $w^1,w^2,...,w^K$ are close to $w^{1^*},w^{2^*},...,w^{k^*}$ by using local data sets $D^1,D^2,...,D^K$ for each client. Moreover, this process needs to ensure that the original data of any client is not exposed to any other client or any third party.

We suppose that there are $N=\sum_{i=1}^{C}=N_i$ data in training, where $N_i$ is the number of $i $-th class ($1 \leq i \leq C$). Each client has $n^k$ sample $(x,y)$ from $p^k$, where $x$ and $y$ denote the input features and corresponding class labels, respectively. 
We define the local loss function of k-th client with the widely used \textit{Cross-Entropy Loss} as:
\begin{equation}\label{eq:ce}
	\mathcal{L}_{ce}^k(w^k)=\sum_{i=1}^{C}p^k(y=i)E_{x|y=i}[logf_i(x;w^k)],
\end{equation}
where $G_i(w^k)=E_{x|y=i}[logf_i(x;w^k)]$ only related to sample classification $i$ and local model $w^k$, the gradient of $G_i(w^k)$ with respect to $w^k$ is $g_i(w^k)=\nabla{G_i(w^k)}$.\\

\noindent\textbf{Contrastive Federated Learning}

In federated learning, each client can only use its dataset for local training. Due to data being Non-IID between different clients, each local model will update towards different local optimal directions. Therefore, there are large gaps between the local models of different clients and they all deviate from the global optimal model. The global model aggregates the local models from the differential client. Thus, its generalization is better than the local model. Based on the idea of contrastive learning, we change the similarity between models by comparing the output of the current local model which is updated in this round, the historical local model which is updated in the previous round, and the global model, thereby controlling the update direction of each local model. Carrying the above insights, we track the challenge of client drift by controlling the update direction of each local model.

Inspired by contrastive learning in siamese network,
we alter the distance between models by changing the cosine similarity between the outputs of the models.  Therefore, we apply the cross-entropy loss function as the basic loss function and introduce a loss function term based on the cosine similarity of the output of the models to correct the updated direction of the local model.
Specifically, we increase the distance between the local models of the previous iteration and this iteration by minimizing the positive cosine similarity between them to mitigate client drift.
At the same time, we decrease the distance between the global and local models in this iteration by minimizing the negative cosine similarity between them to bridge the model differences between each client.

In the local training, each client $C^k$ receives the global model $w_t^f$ and updates the local model $w_t^k$ with its local dataset. To achieve the above objectives, we extract the representation of $x$ from the local model of the last iteration $w_{t-1}^k$ (i.e.,$z^k_{t-1}=R_{w_{t-1}^k}(x)$), the local model of this iteration  $w_t^k$ (i.e.,$z^k_t=R_{w_t^k}(x)$) and the global model $w_t^f$ (i.e.,$z^f_{t}=R_{w_{t}^f}(x)$). Then, we modify the distance of the corresponding model by controlling the cosine similarity of those representations defined as follows.\\

\begin{myDef}
	\label{defpcos}
	The \textbf{positive cosine similarity} between  $z^k_{t-1}$ extracted from $w_{t-1}^k$ and $z^k_t$ extracted from $w_t^k$ as:
\end{myDef}
\begin{equation}\label{pcos}
	\begin{split}
		D(z^k_{t-1},z^k_t)
		=\frac{z^k_{t-1}}{||z^k_{t-1}||}\cdot{\frac{z^k_t}{||z^k_t||}	}.
	\end{split}
\end{equation}
Therefore, the smaller the positive cosine similarity between $z^k_{t-1}$ and $z^k_t$, the farther the distance between  $w^k_{t-1}$ and $w^k_t$, to alleviate client drift.\\

\begin{myDef}  
	\label{defncos}
	The \textbf{negative cosine similarity} between $z^f_t$ extracted from $w^f_t$ and $z^k_t$ extracted from $w^k_t$ as:
\end{myDef}
\begin{equation}\label{ncos}	
	D(z^f_t,z^k_t)
	=-\frac{z^f_t}{||z^f_t||}\cdot{\frac{z^k_t}{||z^k_t||}}.	
\end{equation}

Therefore, the smaller the negative cosine similarity between $z^f_t$ and $z^k_t$, the farther the distance between  $w^f_t$ and $w^k_t$, to bridge the model difference between each client.

We introduce a symmetrized loss function term to the loss function  as:
\begin{equation}\label{loss1}
	\mathcal{L}(w_t^k)=\mathcal{L}_{ce}^k(w^k_t)+\mu[\frac{1}{2}D(z^k_{t-1},z^k_t)+\frac{1}{2}D(z^f_t,z^k_t)],	
\end{equation}
where $\mu$ is a hyperparameter to control the weight of the symmetrized loss function term.\\

\noindent\textbf{Stop-Gradient}

As illustrated in the above method, we can modify the update direction of the local model by controlling the cosine similarity of representations. 
However, the local optimal model is not equivalent to the global model. Thus, the local training may excessively increase the distance between two local models coming from the two iterations and extremely bridge the gap between the local and global model when modifying the update direction of the local model. 
Since the local model over-learns the generalization of the global model, it cannot perform well on the local dataset.

\begin{figure}
	\centering
	\includegraphics[width=0.7\linewidth]{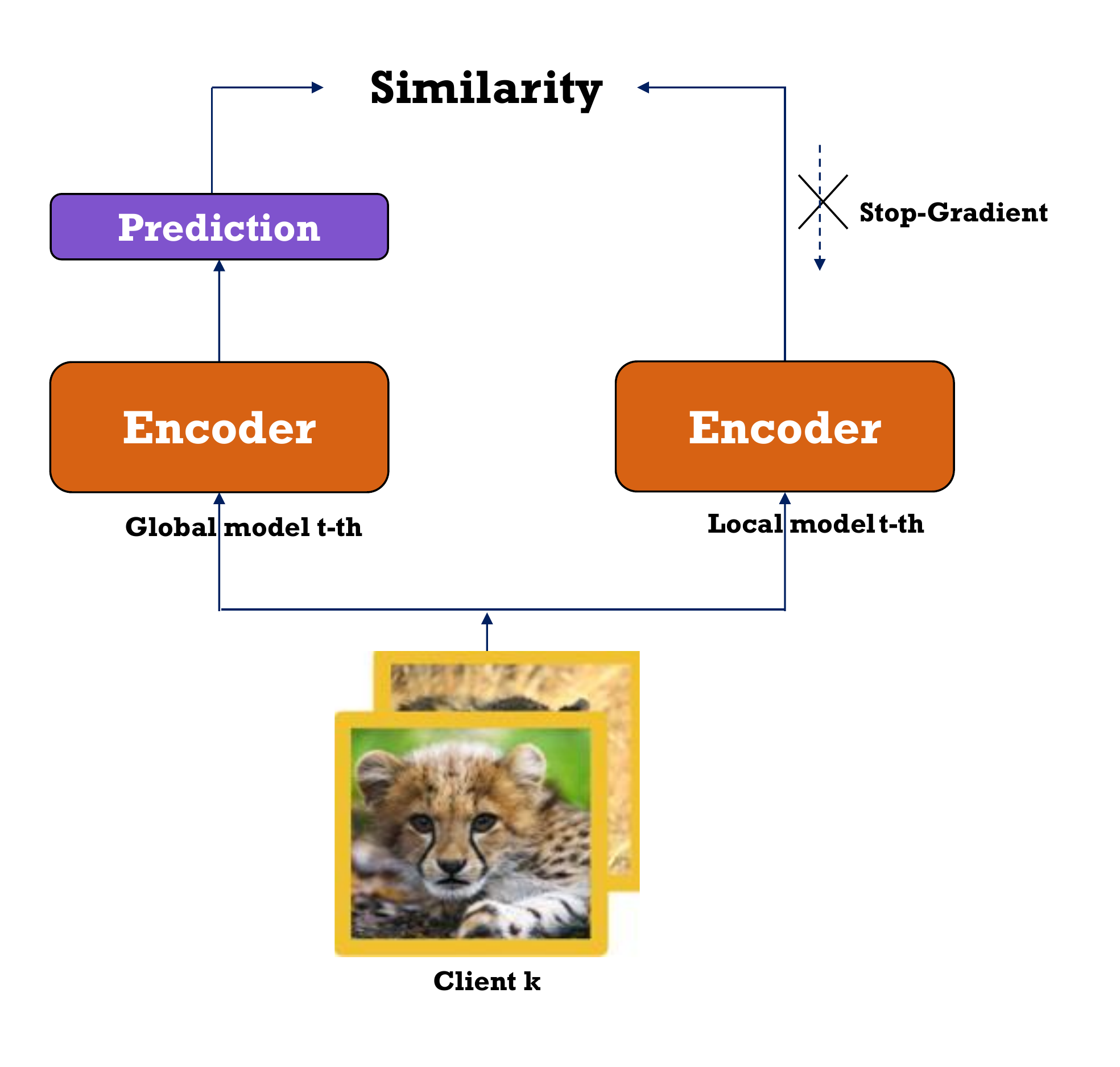}
	\caption{Inputting data to both global and local models. Then, the predicted MLP is applied on the global (or local) model and the stop gradient operation is applied on the local (or global) model. This architecture maximizes the similarity between the two models.}
	\label{fig:simsiam}
\end{figure}

To perform personalized federated learning better, inspired by the \textbf{stop-grad} operation in the loss function of \textbf{SimSiam}, we introduce the \textbf{stop-grad} operation into the symmetrized loss function term to redefine the cosine similarity between the local and the global model, as shown in fig.\ref{fig:simsiam}. 
We assume that there are $M$ local updates between two iterations. After each client receives the global model $w_t^f$  broadcast by the server, the local model $w_{Mt}^k$ updates $M$ epochs, and obtains $w_{Mt+1}^k$,..,$w_{Mt+m}^k$,...,$w_{M(t+1)}^k$. Then, each client uploads the updated local model $w_{M(t+1)}^k$ to the server for aggregation to obtain the new global model $w_{t+1}^f$.
Considering that the distance between the local and global optimal model will get farther with the increase in the number of local updates. Therefore, we personalize the model at each local update.

Specifically, after extracting the representation of each input $x$, our architecture takes a  prediction MLP  head $h$ to transform the representation of one side and match it with the representation of the other side. We denote the output vectors on two sides as $p_{w^{f'}_{t+m}}=h_{w^{f'}_{t+m}}(R_{w_{t+m}^{f'}}(x))$ and $z_{Mt+m}^k=R_{w_{Mt+m}^k}(x)$ and  minimize the negative cosine similarity between them:
\begin{equation}\label{stopgrad-1}
	D(p_{w^{f'}_{t+m}},stopgrad(z_{Mt+m}^k)),
\end{equation}
where $w^{f'}_{t+m}$ is the global model participating in the $m$-th local update. This means that the updated gradient is not back to $R_{w_{Mt+m}^k}$, and $z_{Mt+m}^k$ is regarded as a constant under the stop-gradient operation. Based on this architecture, we rewrite the symmetrized loss function between the current local model $w_{Mt+m}^k$ and global model $w^{f'}_{t+m}$ as:
\begin{equation}\label{stopgrad}
	\begin{aligned}
		\mathcal{L}_{stop}&(w^k_{Mt+m};w^{f'}_{t+m};x)\\&=\frac{1}{2}D(p_{w^{f'}_{t+m}},stopgrad(z_{Mt+m}^k))\\&+\frac{1}{2} D(p_{w^k_{Mt+m}},stopgrad(z_{t+m}^{f'})),
	\end{aligned}		
\end{equation}
where $p_{w^k_{Mt+m}}=h_{w^k_{Mt+m}}(z_{w^k_{Mt+m}})$ and $z_{t+m}^{f'}=R_{w^{f'}_{t+m}}(x)$. In this setting,   $R_{w^k_{Mt+m}}(x)$ can not receive the gradient from the $z_{w^k_{Mt+m}}$ in the first term, but it can receive the gradient from $p_{w^k_{Mt+m}}$ in the second term. 

The presence of stop-gradient implicitly introduces adversarial training between the local and global model in $m$-th local update. The symmetrical loss function alternately optimizes the two models:
\begin{equation}\label{global model}
	\underset{w^{f'}}{argmin}\mathcal{L}(w^{f'}_{t+m},w^k_{Mt+m})\to w^{f'}_{t+m+1},
\end{equation}

\begin{equation}\label{local model}
	\underset{w^k}{argmin}\mathcal{L}(w^{f'}_{t+m+1},w^k_{Mt+m})\to w^k_{Mt+m+1}.
\end{equation}
\textbf{\textit{Training for $w^{f'}$}}: We use SGD to solve the sub-problem (\ref{global model}) and minimize the negative cosine similarity between $p_{w^k_{Mt+m}}$ and $z_{Mt+m}^k$. Then, we use backpropagation to transmit the gradient to the global model $w^{f'}_{t+m}$ for updating. Here $z^k_{Mt+m}$ is regarded as a natural consequence to increase the similarity between global and local models and does not update under the stop-gradient operation.

\noindent\textbf{\textit{Training for $w^k$}}: Similar to sub-problem (\ref{global model}), we minimize the negative cosine similarity between $p_{w^k_{Mt+m}}$ and $z_{t+m}^{f'}$ by sub-problem (\ref{local model}) and back the gradient to the local model $w^k_{Mt+m}$ for controlling the update direction of the local model.

In the same way, we redefined the positive cosine similarity between history local model $w^k_{Mt+m-1}$ and  current local model $w^k_{Mt+m}$ in $m$-th local update:
\begin{equation}\label{hist}
	\begin{aligned}
		\mathcal{L}_{hist}&(w^k_{Mt+m-1};w^k_{Mt+m};x)\\&=D(stopgrad(z^k_{Mt+m-1}),z^k_{Mt+m}).
	\end{aligned}
\end{equation}
Here $z^k_{Mt+m-1}$ is treated as a constant to increase the distance between the current local model $w^k_{Mt+m}$ and the historical local model $w^k_{Mt+m-1}$.

Carrying the above insight, we formulate the loss function of $m$-th local update as:
\begin{equation}\label{m-local loss}
	\begin{aligned}
		\mathcal{L}^k_m(w^k_{Mt+m})=&\mathcal{L}_{ce}(w^k_{Mt+m};x)\\&+\mu (\mathcal{L}_{hist}(w^k_{Mt+m-1};w^k_{Mt+m};x)\\&+ \mathcal{L}_{stop}(w^k_{Mt+m};w^{f'}_{t+m};x))
	\end{aligned}
\end{equation}
Thus, the local objective is to minimize
\begin{equation}\label{local loss}
	\underset{w^k_{Mt}}{min}\mathcal{L}^k(w^k_{Mt})=\frac{1}{M}\sum_{m=1}^{M}\mathcal{L}^k_{m}(w^k_{Mt+m}).
\end{equation}

\subsection{Dual-aggregated}
The above method modifies the update direction of each local model to reduce client drift. Based on the fact that the client uses cosine similarity to change the similarity of models, we further focus on optimizing the global model to improve the accuracy of the federated learning model. Global aggregation usually uses the method of averaging the received local models to generate the global model. Due to the data of each client being Non-IID, the contribution of each local model to the federated learning system is also different.  
Therefore, the global model obtained by averaging each local model can not fit each client dataset well, which brings a fairness problem for the global model.

We propose a novel dual-aggregated approach that two aggregations are performed on the server to personalize the global model. Firstly, the aggregation is consistent with FedAvg, the weighted average of the received local models $w_t^k$ to generate the first global model $w_t^{f_0}$ of the $t$-th iteration on the server as:
\begin{equation}\label{avg}
	w_t^{f_0}=\frac{1}{K}\sum_{k=1}^{K}w_t^k.	
\end{equation}
Secondly, we calculate the cosine similarity between $k$-th local model $w_t^k$ and the first global model $w_t^{f_0}$
\begin{equation}\label{coskf0}
	D(w_t^k,w_t^{f_0})=\frac{w_t^k}{||w_t^k||}\cdot{\frac{w_t^{f_0}}{||w_t^{f_0}||}}.
\end{equation}
Then, taking the ratio of the cosine similarity between the $k$-th local model $w_t^k$ and the global model $w_t^{f_0}$ to the sum of the cosine similarity between each local model and the global model as the weight $\xi_t^k$ of $k$-th local model in the second aggregation:
\begin{equation}\label{weight}
	\xi_t^k=\frac{D(w_t^k,w_t^{f_0})}{\sum_{k=1}^{K}D(w_t^k,w_t^{f_0})}.
\end{equation}
Therefore, the process of the second global aggregation as:
\begin{equation}\label{dual}
	w_t^f=\sum_{k=1}^{K}\xi_t^k w_t^k.
\end{equation}
In the second aggregation, we set a dynamic weight $\xi_t^k$ for each local model parameter according to the cosine similarity between the first global model and local models.
Between the local and global models, the higher the similarity the closer distance, which local model has relatively better generalization and the greater the weight is when aggregated.

\section{Experiments}
In this section, we empirically verify the effectiveness of FedSiam-DA. We summarize the implementation details in Section \ref{4.1} and compare FedSiam-DA with several other federated learning optimization algorithms in Section \ref{4.2}. 
\subsection{Implementation Details}\label{4.1}


\noindent\textbf{Baselines:}
We compare FedSiam-DA with other algorithm including MOON\cite{li2021model},  Fedprox\cite{sahu2018convergence}, and FedAvg\cite{li2019convergence}.\\

\noindent\textbf{Datasets:}
We test the effectiveness of FedSiam-DA using the CIFAR-10 and CIFAR-100 datasets to construct heterogeneous data for each client. Similar to previous studies \cite{li2021model,he2020fedml}, we use a Dirichlet distribution \textbf{Dir($\beta$)} to set Non-IID data across clients. We use the parameter $\beta$ to control the level of data heterogeneity of each client. During implementation, we set $\beta  = 0.3$ and $\beta = 0.5$, with a smaller $\beta$ representing higher data heterogeneity.\\

\noindent\textbf{Network Architecture:}
For both CIFAR-10 and CIFAR-100, we employ ResNet-50 \cite{he2016deep} as the basic backbone. The projection MLP has $2-layers$, and the dimension of output is $z = 256$. The prediction MLP  has BN applied to its hidden fully-connected layers. Its output fully connected does not have BN or ReLU. This prediction MLP head has $2-layers$.\\

\noindent\textbf{Hyperparameters:}
We use SGD\cite{goyal2017accurate,loshchilov2016sgdr} to minimize the loss function and set the learning rate $lr=0.1$. At the same time, the SGD momentum is set to 0.9 and the SGD weight decay is set to $0.00001$. 
The batch size is set to $64$. The number of communication rounds is set to $100$ for CIFAR-10/100 and $20$ for Tiny-ImageNet. Except for special illustrations, the default number of local updates is $5$.
For FedSiam-DA, we tune $\mu$ from ${0.05, 0.1, 0.2, 1}$ and report the
best result. The best $\mu$ of FedSiam-DA for CIFAR-10 and CIFAR-100 is $0.1$.
The best $\mu$ of MOON and FedProx for CIFAR-10 and CIFAR-100 are $1$, $0.01$.

\subsection{Performance Comparison}\label{4.2}
\noindent\textbf{Test Accuracy}

\subsubsection{Client}
\begin{figure}
	\centering
	\includegraphics[width=0.7\linewidth]{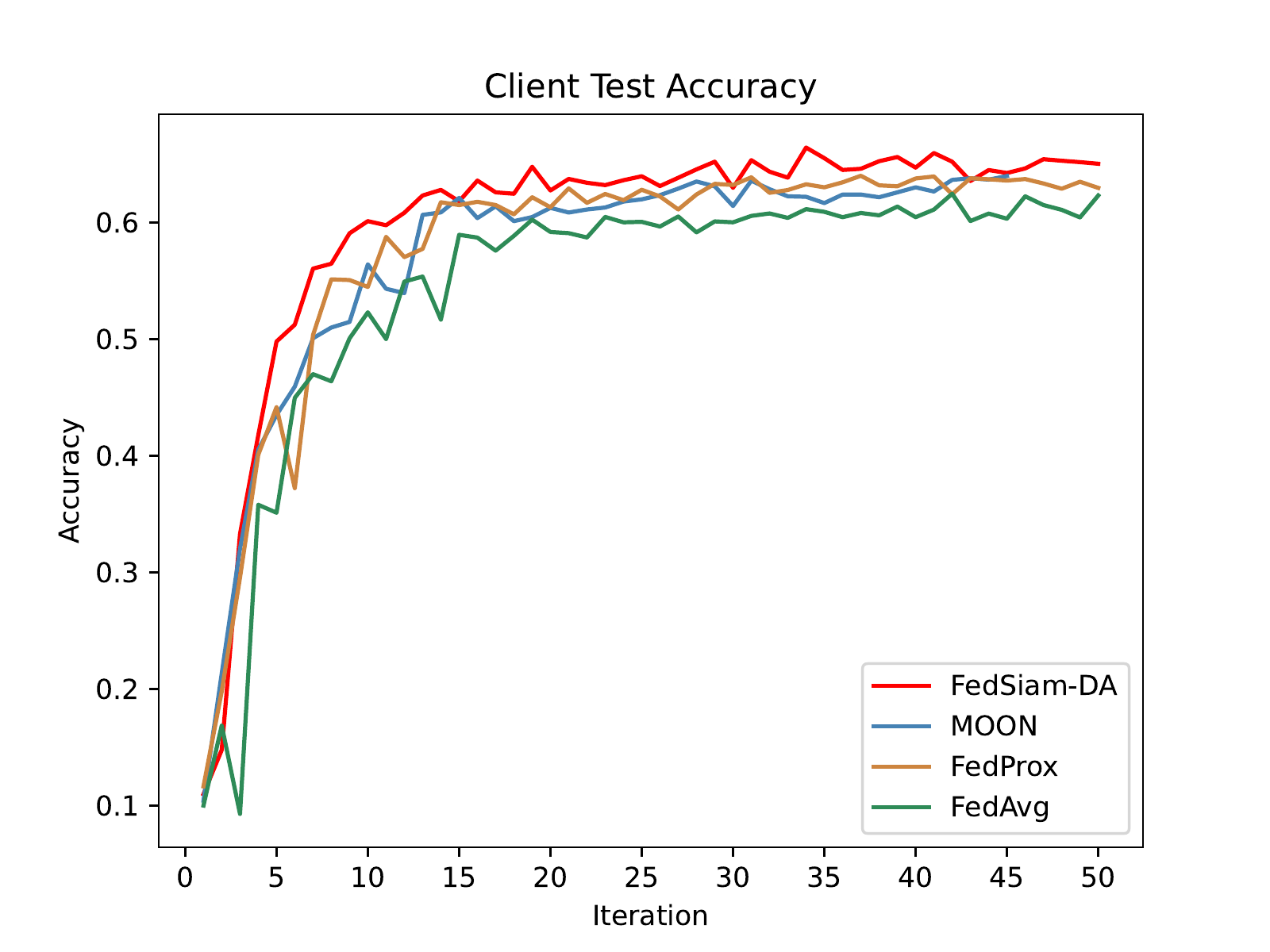}
	\caption{The client test accuracy on CIFAR-100 with 50 communication rounds, FedSiam-AD:$\mu$=0.1, MOON:$\mu$=1, FedProx:$\mu$=0.01.}
	\label{fig:clientacc}
\end{figure}

Fig.\ref{fig:clientacc} shows the client accuracy after we introduce a stop-gradient in the local epoch during local training. We set the best optimal parameters $\mu$ for the loss function term of each algorithm. In the beginning, the speed of FedSiam-DA accuracy improvement is almost the same as MOON and FedProx. However, with the increase in update times, FedSiam-DA can achieve a better accuracy benefit from adversarial training by stop-gradient.

\subsubsection{Server}
\begin{figure}
	\centering
	\includegraphics[width=0.7\linewidth]{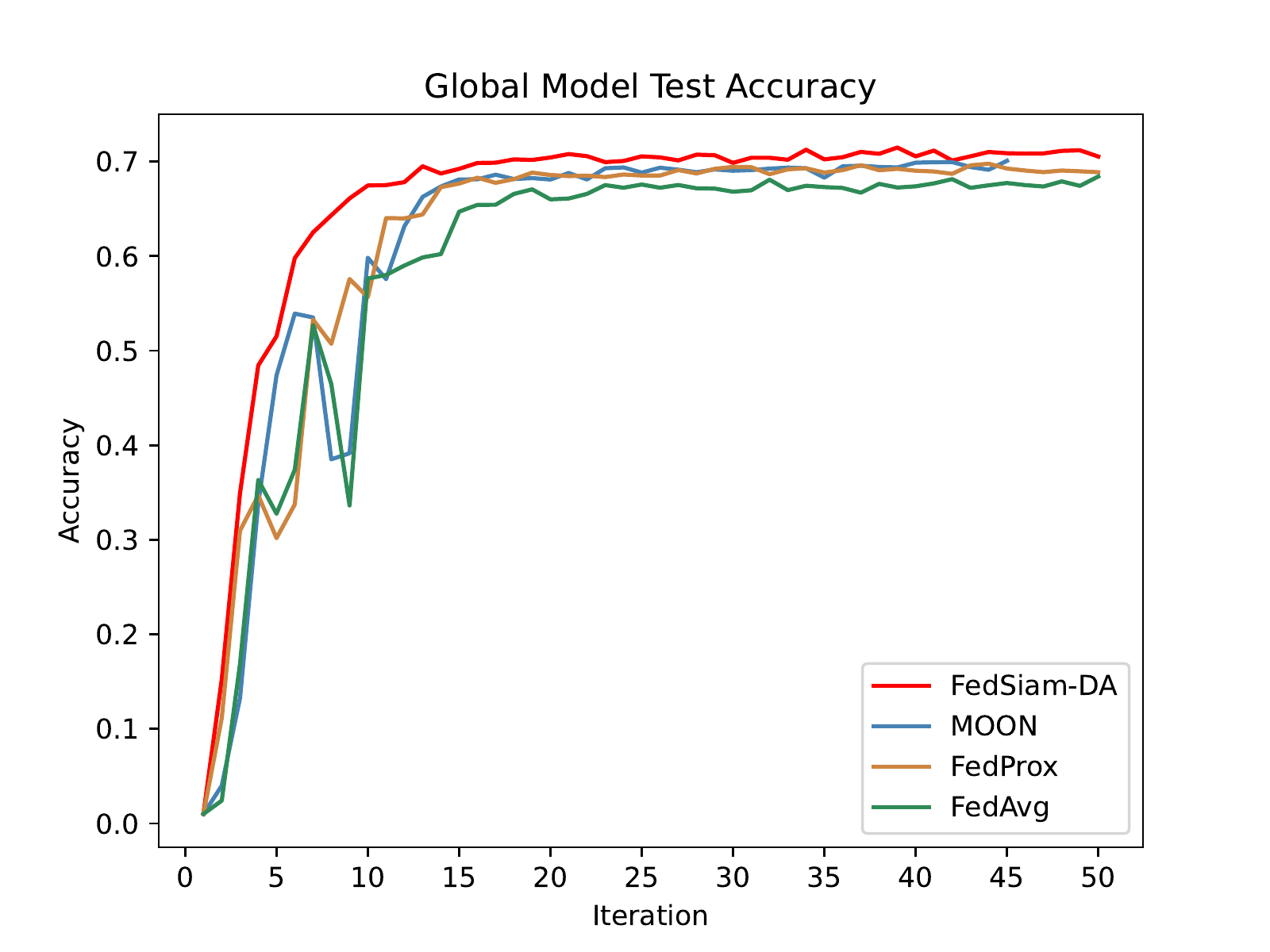}
	\caption[Global Model Acc]{The global model test accuracy on CIFAR-100 with 50 communication rounds, FedSiam-AD:$\mu$=0.1, MOON:$\mu$=1, FedProx:$\mu$=0.01.}
	\label{fig:globalacc}
\end{figure}

\begin{figure}
	\centering
	\includegraphics[width=0.7\linewidth]{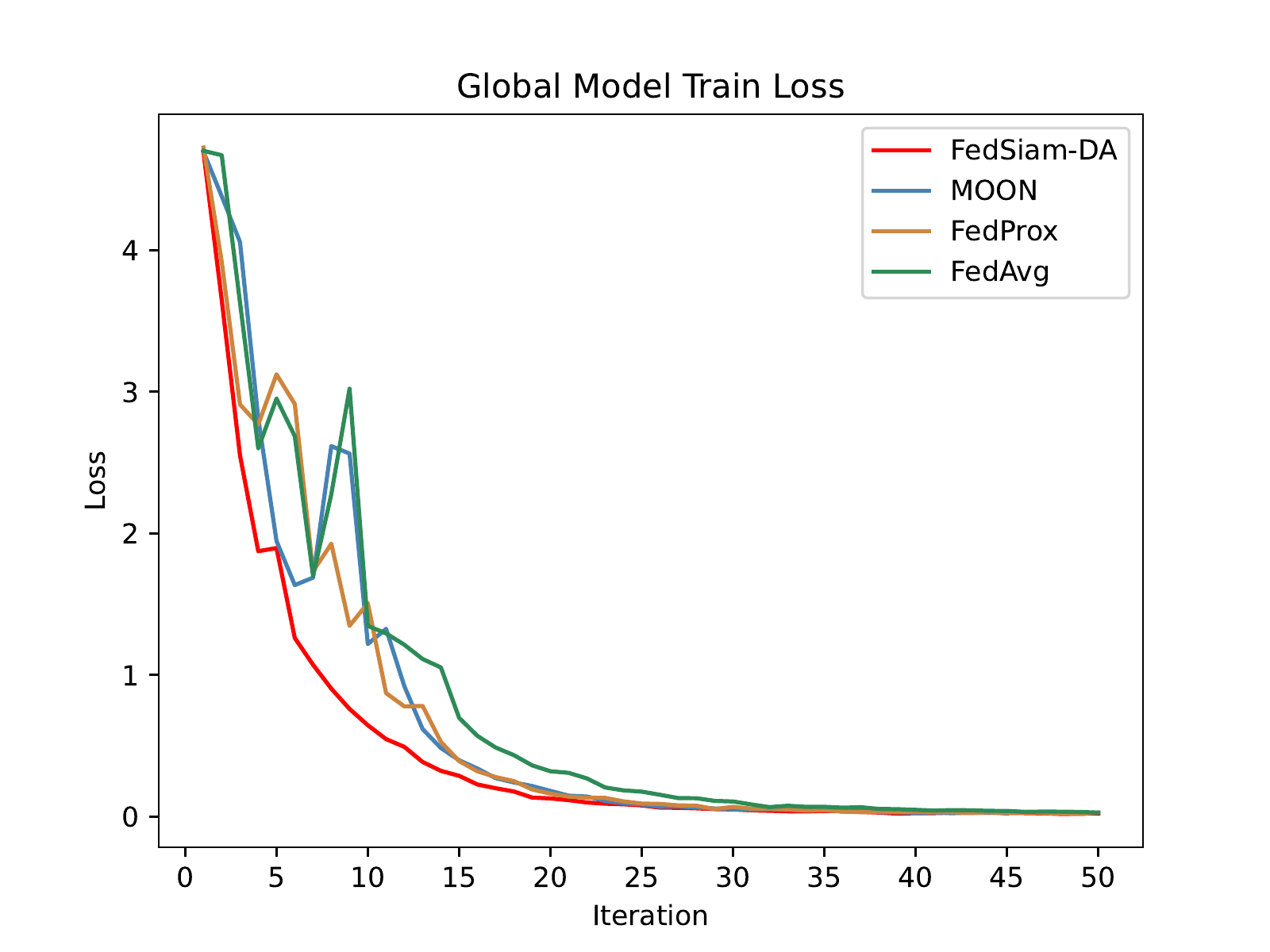}
	\caption[Global Model Loss]{The global model loss function on CIFAR-100 with 50 communication rounds, FedSiam-AD:$\mu$=0.1, MOON:$\mu$=1, FedProx:$\mu$=0.01.}
	\label{fig:globalloss}
\end{figure}

Fig.\ref{fig:globalacc} and fig.\ref{fig:globalloss} show the global model accuracy after we used the dual-aggregated in each round during training.  In the same light, We set the best optimal parameters $\mu$ for the loss function term of each algorithm. 
As we can see, similar to fig.\ref{fig:clientacc}, the FedSiam-DA global model accuracy improves at almost the same rate as other algorithms. However, as the number of iterations increases, the local model continues to personalize. Therefore, the dynamic weight continuously adjusts the proportion of the local model in the global model according to the cosine similarity between the local and global models, and the ability of the global model to learn the knowledge of each local model under the dual aggregation mechanism is continuously improved. In addition, we can see that the loss function obtained by FedSiam-DA is less volatile and smoother than other algorithms.\\

\section{Conclusion}
In this letter, we tackle the challenging problem of personalized federated learning in heterogeneous environments. We introduce a stop-gradient mechanism to adjust the similarity between models more reasonably. The local model can learn the generalization of the global model and perform well on its local dataset. Furthermore, FedSiam-AD uses the dual-aggregated mechanism to personalize the global model on the server to enhance the convergence rate. Our experiments show that FedSiam-AD achieves significant improvement over other algorithms in heterogeneous environments.

\bibliographystyle{IEEEtran}        

\bibliography{reference}

\end{document}